# Weakly-Supervised Convolutional Neural Networks for Multimodal Image Registration

Yipeng Hu[*1,2], Marc Modat[1,3], Eli Gibson[1], Wenqi Li[1,3], Nooshin Ghavami[1], Ester Bonmati[1], Guotai Wang[1,3], Steven Bandula[4], Caroline M. Moore[5], Mark Emberton[5], Sébastien Ourselin[1,3], J. Alison Noble[2], Dean C. Barratt[**1,3], Tom Vercauteren[**1,3]

[1] Centre for Medical Image Computing, Department of Medical Physics and Biomedical Engineering, University College London, London, UK
[2] Institute of Biomedical Engineering, Department of Engineering Science, University of Oxford, Oxford, UK
[3] Wellcome / EPSRC Centre for Interventional and Surgical Sciences, University College London, London, UK
[4] Centre for Medical Imaging, University College London, London, UK
[5] Division of Surgery and Interventional Science, University College London, London, UK
*Corresponding Email: yipeng.hu@ucl.ac.uk
** Equal contribution

**Abstract**: One of the fundamental challenges in supervised learning for multimodal image registration is the lack of ground-truth for voxel-level spatial correspondence. This work describes a method to infer voxel-level transformation from higher-level correspondence information contained in anatomical labels. We argue that such labels are more reliable and practical to obtain for reference sets of image pairs than voxel-level correspondence. Typical anatomical labels of interest may include solid organs, vessels, ducts, structure boundaries and other subject-specific *ad hoc* landmarks. The proposed end-to-end convolutional neural network approach aims to predict displacement fields to align multiple labelled corresponding structures for individual image pairs during the training, while *only* unlabelled image pairs are used as the network input for inference. We highlight the versatility of the proposed strategy, for training, utilising diverse types of anatomical labels, which need not to be identifiable over all training image pairs. At inference, the resulting 3D deformable image registration algorithm runs in real-time and is fully-automated without requiring any anatomical labels or initialisation. Several network architecture variants are compared for registering T2-weighted magnetic resonance images and 3D transrectal ultrasound images from prostate cancer patients. A median target registration error of 3.6 mm on landmark centroids and a median Dice of 0.87 on prostate glands are achieved from cross-validation experiments, in which 108 pairs of multimodal images from 76 patients were tested with high-quality anatomical labels.

## 1    Introduction

Multimodal image registration aims to spatially align medical images produced from different imaging modalities. Among many other medical imaging applications, this is useful in minimally- or none-invasive image-guided procedures, in which a common strategy is to fuse the detailed diagnostic information from quality pre-procedural images with intra-procedural imaging that is typically restricted by the interventional requirements, such as portability, accessibility, temporal resolution, limited field of view and controlled dosage for contrast agent or radiation.

Classical pairwise intensity-based image registration methods are in general based on optimising image similarity, a metric indicating how well image intensities correspond (Hill et al., 2001). However, in many interventional applications, engineering a multimodal similarity metric that is sufficiently robust for clinical use is challenging. Potential difficulties include: 1) different physical acquisition processes may generate statistical correlation between imaging structures that do not correspond to the same anatomical structures, violating one of the underlying assumptions for most intensity-based similarity measures (Zöllei et al., 2003); 2) the spatial and temporal variabilities in the intra-procedural imaging, partly due to user-dependency (Noble, 2016), is complex to summarise with simple statistical properties or information-theory-based measures; and 3) intraoperative time constraints prevent the use of better imaging quality as it typically requires significant imaging or processing time, as well as the use of computationally-intensive approaches, such as exhaustive global optimisation.

Alternative feature-based image registration methods, when features are extracted automatically, face similar challenges. Manual anatomical feature selection for registration is user-dependent and often costly or even infeasible intraoperatively but arguably remains the most robust method for multimodal image registration for many intra-procedural applications (Viergever et al., 2016). Semi-automated or assisted medical image segmentation is a promising research direction to support registration (Wang et al., 2017), but it has not yet demonstrated clinical value in fast evolving interventional applications.

In this work, we focus on one exemplar application of interventional multimodal image registration which is to register pre-procedural multi-parametric magnetic resonance (MR) images to intra-procedural transrectal ultrasound (TRUS) images for prostate cancer patients (Pinto et al., 2011; Rastinehad et al., 2014; Siddiqui et al., 2015). Multi-parametric MR imaging (Dickinson et al., 2011), including recent development of hyperpolarised imaging (Wilson and Kurhanewicz, 2014) and computational methods based on diffusion-weighted imaging (Panagiotaki et al., 2015), have shown favourable results in diagnosing and staging prostate cancer. This has already been recommended to form a part of a standard clinical pathway in some countries, including the UK (Vargas et al., 2016). On the intra-procedural side, TRUS imaging is routinely used for guiding the majority of targeted biopsies and focal therapies, but it provides limited value in differentiating cancerous tissue from healthy surroundings. Fusing the MR and TRUS images, can enable accurate detection, localisation and treatment of low- to medium-risk disease in TRUS-guided procedures (Valerio et al., 2015). However, like most other ultrasound-guided medical procedures, this represents a typical example where no robust image similarity measure has been demonstrated. For example, anatomically different imaging structures, such as the prostate inner-outer gland separation, a cleavage plane known as the surgical capsule, defined on TRUS images (Ethan J. Halpern, 2008) and the central-peripheral zonal boundary visible on MR images, appear as being similar in the two types of images and thus possess strong statistical correlation between them. This leads to false alignment using most, if not all, of the established intensity-based similarity measures and the associated registration methodologies, such as the work by Rueckert et al. (Rueckert et al., 1999).

To alleviate some of the aforementioned problems from both the intensity- and feature-based methods in registration applications of this type, a class of model-to-image fusion methods have been proposed (Hu et al., 2012; Khallaghi et al., 2015; van de Ven et al., 2015; Wang et al., 2016a), in which motion models of the prostate glands obtained from MR image are aligned to the surface of the gland capsule automatically or semi-automatically. These methods suffer from two limitations. First, the subject-specific pairwise registration requires correspondent features to be extracted from both images. We previously argued that the only common features of the prostate gland that are consistently available from both images are the capsule surface while *ad hoc* landmarks can be found on a case-by-case basis for validation purpose (Hu, 2013). Indeed, the gland boundary has been the feature of interest to match in most of these mentioned algorithms. Second, partly as a result of the availability of the sparse features, some form of a motion prior is required to regularise the non-rigid registration methods (De Silva et al., 2017; Hu et al., 2015, 2011; Khallaghi et al., 2015; Wang et al., 2016b). The learning of the motion models is highly application-dependent and usually not generalisable to other medical applications or different imaging protocols for the same application, such as pathological cases or interventions with different surgical instruments.

Supervised representation learning (Bengio et al., 2013), especially methods using convolutional neural networks (LeCun et al., 2015, 1998), has the potential to optimise medical image representation in a regression network that predicts spatial correspondence between a pair of given images, without human-engineered image features or intensity-based similarity measures. However, voxel-level ground-truth for learning correspondence are scarce and, in most scenarios, impossible to reliably obtain from medical image data. Alternative methods to learn similarity measures, e.g. (Simonovsky et al., 2016), also require non-trivial ground-truth labels and, to our best knowledge, have not been proposed for registering MR and ultrasound images. Several methods have been proposed to procure large numbers of pseudo-ground-truth transformations for training, such as those from simulations (Krebs et al., 2017; Miao et al., 2016; Sokooti et al., 2017), existing registration methods (Rohé et al., 2017) or manual rigid alignment (Liao et al., 2017). Recently-proposed machine-learning-based image registration methods have relied on image-similarity-driven unsupervised learning (Cao et al., 2017; de Vos et al., 2017; Wu et al., 2013; Yang et al., 2017), meaning that these methods inherit the key shortcomings of classical intensity-based image registration algorithms.

We argue that higher-level corresponding structures are much more practical to annotate reliably with anatomical knowledge. Such labels can be used to highlight in pairs of images the same organs and boundaries between them, pathological regions, and other anatomical structures, morphological or physiological features appearing in both images, and can serve as weak labels for training the prediction of lower-level voxel correspondence. Moreover, subject-specific landmarks that are only inconsistently available from all image pairs may also contribute to finding detailed voxel correspondence, especially from interventional data. For instance, spatial distributions of calcification scatters and water-based cysts are highly patient-specific (see an example in Fig.1). Although readily identifiable in many pairs, they have mostly been used for validation purposes (Hu et al., 2012; van de Ven et al., 2013; Wang et al., 2016a). In this work, we introduce a novel framework which uses these anatomical labels and full image voxel intensities as training data, to enable a fully-automatic, deformable image registration that requires only unlabelled image data during inference.

Initial results were reported in an abstract on our preliminary work (Hu et al., 2018). We summarise the substantially extended contributions contained in this paper: 1) a detailed methodology description for the weakly-supervised image registration framework is presented in Section 2.1; 2) a new efficient multiscale Dice for weakly-supervised registration network training is described in Section 2.2; 3) a novel memory-efficient network architecture is proposed without using the previously proposed global affine sub-network in Section 2.3; and 4) rigorous analysis comparing different network variations and classical pairwise registration algorithms are reported in Section 4 and significantly improved results are also presented.

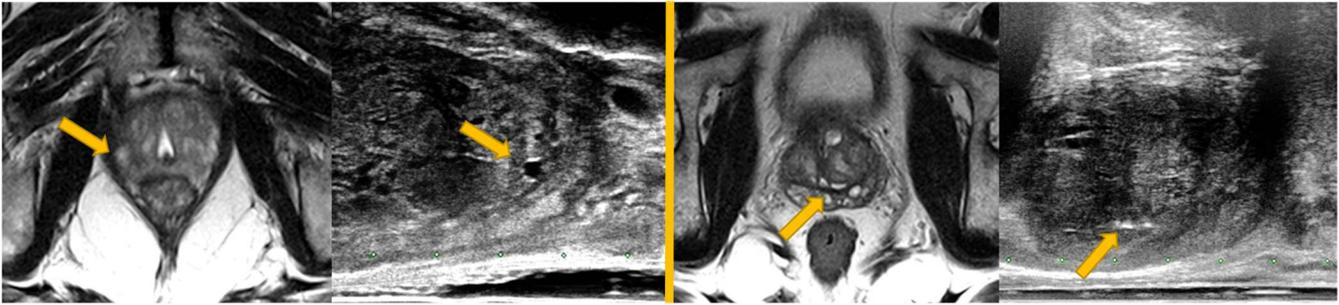

**Fig.1.** Examples of corresponding training landmark pairs used in this study, a water-filled cyst (on the left MR-TRUS image pair) and a cluster of calcification deposit (on the right image pair). These *ad hoc* landmarks are not consistently available for all patient data and have usually been identified only for validation purpose in previous studies. Details are discussed in Section 1 and the network training utilising these landmarks is described in Section 2.

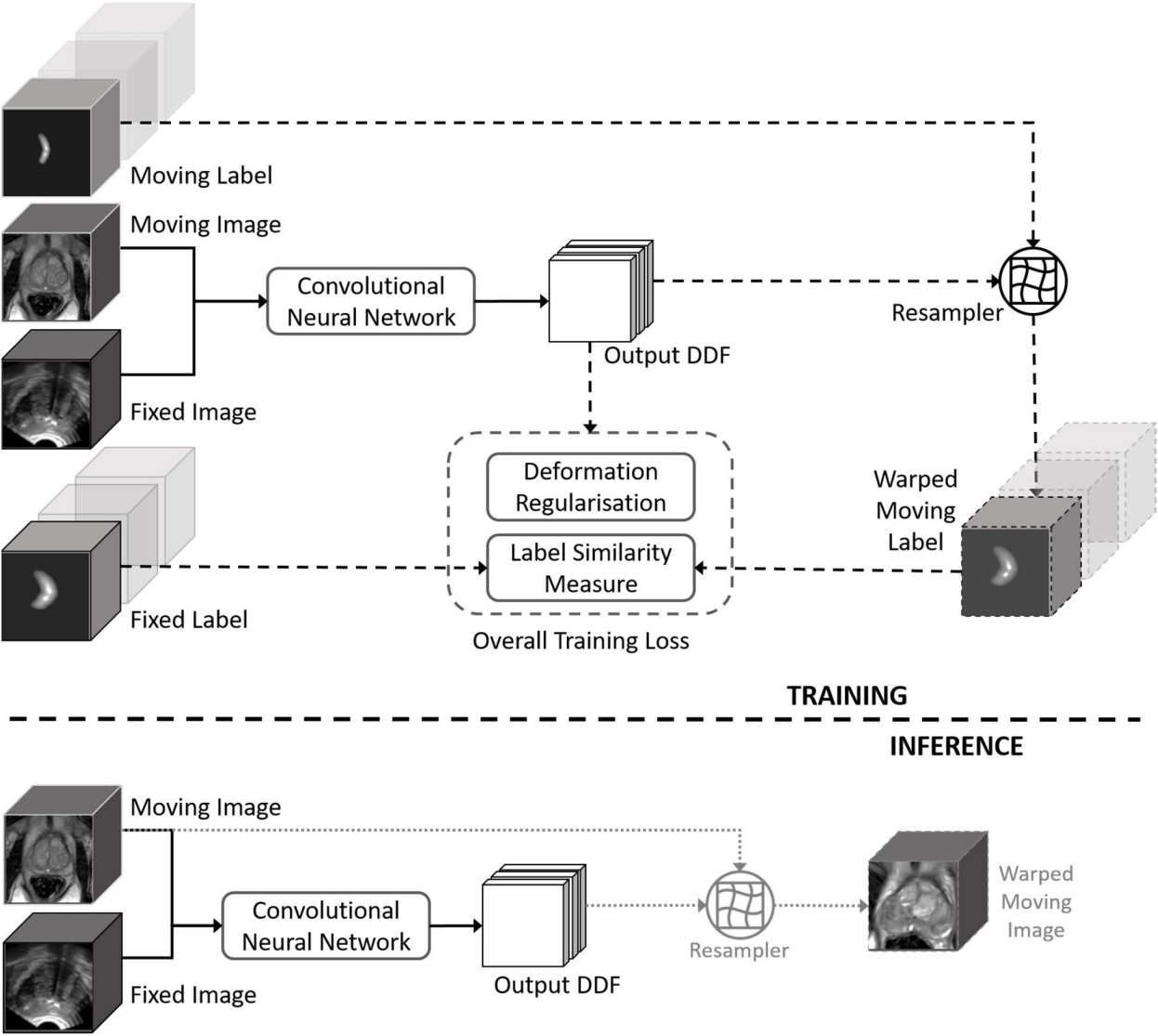

**Fig.2.** The upper part illustrates the training strategy of the proposed weakly-supervised registration framework (described in Section 2.1), where the dashed lines indicates data flows only required in training. The lower part depicts the resulting inference (indicated by the solid lines), i.e. registration predicting the output DDF, requiring only the image pair, with which the moving image may be warped to align with the fixed image (dotted lines).

## 2 Method

### 2.1 A Weakly-Supervised Image Registration Framework

Given $N$ pairs of training moving- and fixed images, $\boldsymbol{x}^A = \{\boldsymbol{x}_n^A\}$ and $\boldsymbol{x}^B = \{\boldsymbol{x}_n^B\}$, respectively, $n = 1, \ldots, N$. On the $n^{th}$ image pair, $M_n$ pairs of moving- and fixed labels $\boldsymbol{l}^A = \{\boldsymbol{l}_{mn}^A\}$ and $\boldsymbol{l}^B = \{\boldsymbol{l}_{mn}^B\}$ represent corresponding regions of anatomy, $m = 1, \ldots, M_n$. We formulate the training of a neural network to predict the voxel correspondence, which is represented by a dense displacement field (DDF) $\boldsymbol{u}_n$, as a weakly-supervised learning problem that maximises a utility function indicating the *expected label similarity* over $N$ training image pairs:

$$J = \frac{1}{N}\sum_{n=1}^{N}\frac{1}{M_n}\sum_{m=1}^{M_n} J_{mn}(\boldsymbol{l}_{mn}^B, \boldsymbol{y}_{mn}^A) \quad (Eq.1)$$

where the inner summation represents the image-level label similarity, averaging a label-level similarity measure over $M_n$ labels associated with the $n^{th}$ image pair. In this work, the label-level similarity is computed between the fixed label $\boldsymbol{l}_{mn}^B$ and the spatially warped moving label $\boldsymbol{y}_{mn}^A = f_T(\boldsymbol{l}_{mn}^A, \boldsymbol{u}_n)$ with the displacements $\boldsymbol{u} = \{\boldsymbol{u}_n(\boldsymbol{x}_n^A, \boldsymbol{x}_n^B, \boldsymbol{\theta})\}$ being predicted by the neural network parameterised by $\boldsymbol{\theta}$, as illustrated in Fig.2. The network training aims to minimise the negative utility function balanced with a deformation regularisation $\Omega(\boldsymbol{u})$ penalising non-smooth displacements, weighted by a hyper-parameter $\alpha$:

$$\hat{\boldsymbol{\theta}} = arg\min_{\boldsymbol{\theta}}[-J(\boldsymbol{x}^A, \boldsymbol{x}^B, \boldsymbol{l}^A, \boldsymbol{l}^B; \boldsymbol{\theta}) + \alpha \cdot \Omega(\boldsymbol{u})] \quad (Eq.2)$$

As motivated in the Introduction, we emphasize that such a loss does not incorporate any intensity-based similarity term which is proven to be unreliable in our application. During training, we use a standard stochastic *K*-minibatch gradient descent optimisation (Goodfellow et al., 2016) which requires an unbiased estimator of the additive *batch gradients* in each minibatch $\frac{\partial J}{\partial \boldsymbol{\theta}} = \frac{1}{K}\sum_{k=1}^{K}\frac{\widehat{\partial J_k}}{\partial \boldsymbol{\theta}}, k = 1, \ldots, K$. To avoid the non-trivial computation of minibatch gradients with a variable number of labels and to simplify the implementation, we propose to construct such a gradient estimator by a two-stage sampling: $K$ image pairs are sampled uniformly in the first stage, then in second stage single label pairs are sampled uniformly from those associated with the previously-sampled image pairs. With this approach, each minibatch contains an equal number of $K$ image-label pairs, from which $\frac{\widehat{\partial J_k}}{\partial \boldsymbol{\theta}}$ is estimated. Given the first-stage-sampled image pairs, let's consider $E_2\left(\frac{\widehat{\partial J_k}}{\partial \boldsymbol{\theta}}\right) = \frac{\partial J_k}{\partial \boldsymbol{\theta}}$ as the conditional expectation of the estimated gradients over the label pairs sampled in the second stage. With the first-stage expectation $E_1[\cdot]$ over image pairs, it can be shown that the minibatch gradients $\frac{\widehat{\partial J}}{\partial \boldsymbol{\theta}}$ computed from the two-stage clustering sampling is unbiased:

$$E\left(\frac{\widehat{\partial J}}{\partial \boldsymbol{\theta}}\right) = E_1\left[E_2\left(\frac{\widehat{\partial J}}{\partial \boldsymbol{\theta}}\right)\right] = E_1\left[E_2\left(\frac{1}{K}\sum_{k=1}^{K}\frac{\widehat{\partial J_k}}{\partial \boldsymbol{\theta}}\right)\right] = E_1\left[\frac{1}{K}\sum_{k=1}^{K}E_2\left(\frac{\widehat{\partial J_k}}{\partial \boldsymbol{\theta}}\right)\right] = E_1\left[\frac{1}{K}\sum_{k=1}^{K}\frac{\partial J_k}{\partial \boldsymbol{\theta}}\right] = \frac{\partial J}{\partial \boldsymbol{\theta}} \quad (Eq.3)$$

We summarise several advantages of the proposed framework illustrated in Fig.2. First, the modality-independent label similarity is computed between the warped moving label and the fixed label, neither of which are used as input to the network. Therefore, they are not required in the inference stage, i.e. actual registration. Second, samples of different types of labels can be fed to the training without requiring consistent number or types of anatomical structures being labelled; and potentially very large number of labels for each image pair can be used without increasing memory usage. Third, the moving and fixed images are the only inputs to the neural network without the need to define an explicit intensity-based image similarity measure that has to be tailored for different modality pairs. Matching intensity patterns will be learned by the network trained to optimise for latent label correspondence. Fourth, different regularisation terms can be added, such as bending energy (Rueckert et al., 1999), *L¹*- or *L²*-norm of the displacement gradients (Fischer and Modersitzki, 2004; Kumar and Dass, 2009; Vishnevskiy et al., 2017), in addition to the network architectural constraints.

*2.2    Multiscale Dice for Measuring Label similarity*

Direct use of classical overlap metrics between binary anatomical labels, such as those based on Dice, Jaccard and cross-entropy, are not appropriate for measuring label similarity in the context of image registration. For example, they do not consider the spatial information when two foreground objects do not overlap. All of them approach extreme values, becoming invariant to the distance between the objects. Our initial work reported to use a cross-entropy with a heuristic label smoothing approach based on re-weighted inverse distance transform (Hu et al., 2018). The warped labels were approximated by interpolating pre-computed label maps, as the distance transform is neither differentiable nor efficient to compute in each iteration.

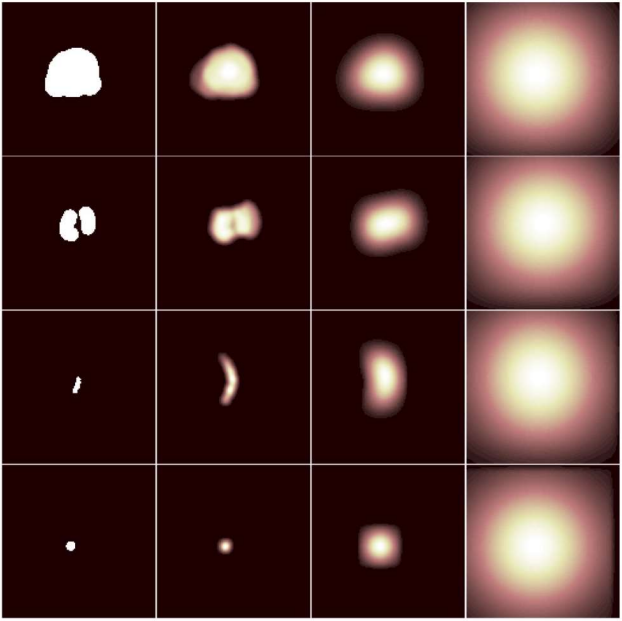

**Fig.3.** Gaussian-based multiscale representation of the example labels used in the proposed label similarity measure. Rows illustrate different types of landmarks, slices from two prostate glands, a urethra and a cyst, from top to bottom; columns are examples of Gaussian smoothed binary labels (first column, $\sigma = 0$) with different standard deviations, $\sigma = 0, 2, 8, 32$ from left to right. The details are described in Section 2.2.

Here, we propose an alternative label similarity measure based on a multiscale Dice. The soft probabilistic Dice (Milletari et al., 2016) $S_{Dice}$ has been shown to be less sensitive to class imbalance in medical image segmentation tasks (Sudre et al., 2017). Between two labels $\mathbf{a} = \{a_i\}$ and $\mathbf{b} = \{b_i\}$, $a_i, b_i \in [0,1]$, $S_{Dice}$ is given as follows:

$$S_{Dice}(\mathbf{a}, \mathbf{b}) = \frac{2 \sum_{i=1}^{I} a_i \cdot b_i}{\sum_{i=1}^{I} a_i + \sum_{i=1}^{I} b_i} \ (Eq.4)$$

where, $i = 1, \ldots, I$, over $I$ image voxels. Given the pair of binary labels $\boldsymbol{l}_k^B = \{(l_k^B)_i\}$ and $\boldsymbol{y}_k^A = \{(y_k^A)_i\}$ in a training minibatch. To better capture spatial information between labels, the proposed multiscale Dice is defined as:

$$J_k = \frac{1}{Z} \sum_\sigma S_{Dice}(f_\sigma(\boldsymbol{l}_k^B), f_\sigma(\boldsymbol{y}_k^A)) \ (Eq.5)$$

where, $f_\sigma$ is a 3D Gaussian filter with an isotropic standard deviation $\sigma$. In this work, the number of scales $Z$ is set to 7, with $\sigma \in \{0, 1, 2, 4, 8, 16, 32\}$ in mm. $f_{\sigma=0}$ is equivalent to filtering with a Dirac delta function, meaning that an unfiltered binary label at original scale is also included when averaging $S_{Dice}$ values. An illustration of the multiscale filtering on the anatomical labels are provided in Fig.3. The proposed Gaussian filtering based multiscale loss metric is differentiable and, if required, can be efficiently evaluated on-the-fly after non-rigid warping and necessary data augmentation.

For comparison, the proposed multiscale approach is also adapted with a classification loss using a negative cross-entropy:

$$S_{CE}(\mathbf{a}, \mathbf{b}) = \sum_{i=1}^{I} \sum_{c=1}^{2} p_c(a_i) \log p_c(b_i) \ (Eq.6)$$

where $p_c$ represents the class probabilities between the foreground- and background classes, $c = \{1, 2\}$. A numerically stable implementation clipping extreme input probabilities can be used in this case.

We summarise several technical considerations in designing the proposed label similarity measure in Eq (5): 1) it has the effect of penalising high confidence binary predictions, similar to the label-smoothing regularisation approaches (Pereyra et al., 2017; Szegedy et al., 2016); 2) from a classification perspective, it further improves the gradient balance between foreground- and background classes over voxel samples in training, as a result of reducing the difference between the expected class probabilities (Lawrence et al., 2012); 3) it provides non-saturating gradients from anatomical labels, especially for those with smaller volumes, due to the high variance spatial smoothing at larger scales; 4) it is highly efficient to compute with recursive and separable convolution kernels.

## 2.3  Network Architecture

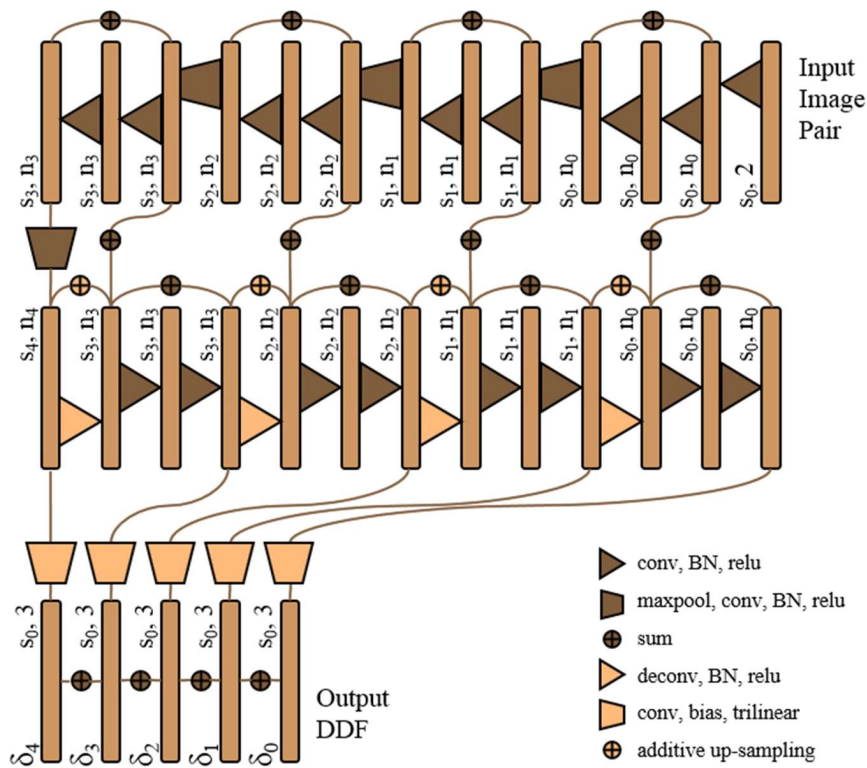

**Fig.4.** Illustration of the proposed registration network architecture.

As shown in our preliminary work (Hu et al., 2018), a global sub-network predicting an affine transformation can be combined with a jointly-trained local sub-network predicting a local DDF, in order to overcome the practical difficulty in propagating the gradients from the deformation regulariser to regions with less supporting label data. In this work, we describe a new architecture utilising a single network to predict displacement summed over different resolution levels. The lower-level displacement summands provide global information, similar to that of the global sub-network but without significant memory usage by the global sub-network. These approaches are compared in Section 3.2.

Following our previous work in segmenting prostate gland from TRUS images (Ghavami et al., 2018) and the prior art for learning optical flow (Ilg et al., 2017), the network is designed as a 3D convolutional neural network with four down-sampling blocks, followed by four up-sampling blocks. As illustrated in Fig.4, the network is more densely connected than the U-Net proposed for image segmentation (Ronneberger et al., 2015) and also has less memory requirement, featuring three types of previously proposed summation-based residual shortcuts, 1) four summation skip layers shortcutting the entire network at different resolution levels (Yu et al., 2017), 2) eight standard residual network shortcuts summing feature maps over two sequential convolution layers (He et al., 2016), and 3) four trilinear additive up-sampling layers are added over the transpose-convolution layers (Wojna et al., 2017).

The benefits of deeper supervision using denser connections have been shown in computer vision tasks (He et al., 2016; Huang et al., 2016; Lee et al., 2015; Szegedy et al., 2015) and medical image analysis (Dou et al., 2017; Garcia-Peraza-Herrera et al., 2017; Gibson et al., 2017a). Besides the thoroughly applied residual shortcuts described above, we introduce summation-based skip layers to the displacement space across different resolution levels $s_{0-4}$. As sketched in the lower part of Fig.4, each side of the up-sampling blocks extends to a node to predict a trilinear-up-sampled displacement summand $\delta_{1-4}$ at levels $s_{1-4}$, after an additional convolution layer added to a bias term, without batch normalisation or standard nonlinear activation. These summands, with the size of the output DDF, are then added to the summand $\delta_0$ at the input image resolution level $s_0$, to predict a single output DDF.

Physically parametrised global transformations such as rigid and affine models are sensitive to network initialisation, as in training spatial transformer networks (Jaderberg, 2015). To a lesser degree, the registration networks predicting displacements suffer the same problem. The design of these summand nodes allows random initialisation with zero mean and a small variation on the convolution weights and bias (on the displacement skip layers) with controlled magnitude of the initial DDFs, such that the warped labels generate meaningful initial gradients. The trilinear sampling provides bounded nonlinear activation between linear convolutions.

The described additive displacement skip layers are more efficient to compute and, potentially, easier to train, comparing to composing displacements at different levels or concatenating warped input images (Ilg et al., 2017; Yu et al., 2016), both requiring resampling. It is noteworthy that the described four displacement skip layers are determined by the network up-sampling levels, therefore are independent to the choice of scales in the label similarity measure above-described in Section 2.2, which evaluates the loss with respect to the single output DDF.

As illustrated in Fig.4, the first feature maps begin with $n_0$ initial channels, successively doubles the number of channels and halves the feature map size with the down-sampling blocks, and *vice versa* with the up-sampling blocks. Each of these blocks consists of two convolution- and batch normalisation (BN) layers with rectified linear units (relu). 3D down- and up-sampling are achieved respectively by max-pooling (maxpool) and transpose-convolution (deconv) layers, both with strides of two. All convolution layers have $3 \times 3 \times 3$ kernels, except for $7 \times 7 \times 7$ kernels used in the first convolution layer to ensure sufficient receptive field.

## 3 Experiments

### 3.1 Data

A total of 108 pairs of T2-weighted MR and TRUS images from 76 patients were acquired during SmartTarget® clinical trials (Donaldson et al., 2017). Each patient had up to three image data sets due to the multiple procedures he entered, i.e. biopsy and therapy, or multiple ultrasound volumes acquired at the beginning and the conclusion of a procedure according to the therapy trial protocol ("SmartTarget: BIOPSY," 2015, "SmartTarget THERAPY," 2014). Using a standard clinical ultrasound machine (HI-VISION Preirus, Hitachi Medical Systems Europe) equipped with a bi-plane (C41L47RP) transperineal probe, a range of 57 - 112 TRUS frames were acquired in each case by rotating a digital transperineal stepper (D&K Technologies GmbH, Barum, Germany) with recorded relative angles covering the majority of the prostate gland. These parasagittal slices were then used to reconstruct a 3D volume in Cartesian coordinates (Hu et al., 2017). Both MR and TRUS images were normalised to zero-mean with unit-variance intensities after being resampled to 0.8×0.8×0.8 mm³ isotropic voxels.

From these patients, a total of 834 pairs of corresponding anatomical landmarks were labelled by two medical imaging research fellows and a research student using an in-house voxel-painting tool on the original image data, and all were verified by second observers including a consultant radiologist and a senior research fellow. Prostate gland segmentations on MR images were acquired as part of the trial protocols (Donaldson et al., 2017). The gland segmentations on TRUS images were manually edited based on automatically contoured prostate glands on original TRUS slices (Ghavami et al., 2018). Besides full gland segmentations for all cases, the landmarks include apex, base, urethra, visible lesions, junctions between the gland, gland zonal separations, vas deference and the seminal vesicles, and other patient-specific point landmarks such as calcifications and fluid-filled cysts (see also Fig.1 and Fig.3 for examples). The label pairs used in this study include 108 (12.9%) pairs of gland segmentations, 213 (25.5%) apex or base pairs, 214 (25.7%) corresponding structures on zonal boundaries, 37 (4.4%) on urethra and 262 (31.4%) patient-specific regions of interest such as calcification sediments and cysts, with an average volume of 0.39±0.21 cm³ and a range of [0.13, 18.0] cm³ excluding the gland segmentations. The landmark annotation process took more than 200 hours. The anatomical labels, represented by binary masks, were resampled to the sizes and resolutions of the associated MR or TRUS images, and were re-grouped for training (described in Section 2.1) and for validation in a cross-validation scheme described in Section 3.3.

### 3.2 Implementation and Network Training

The described methods were implemented in TensorFlow™ (Abadi et al., 2016) with a trilinear resampler module and a 3D image augmentation layer adapted from open-source code in NiftyNet (Gibson et al., 2017b). Re-implementation of all the networks reported in the experiment are available as part of NiftyNet (niftynet.io). Each image-label pair was transformed by a random affine transformation *without* flipping before each training iteration for data augmentation. Each network was trained with a 12GB NVIDIA® Pascal™ TITAN Xp general-purpose graphic process unit (GPU) for 48 hours on a high-performance computing cluster.

#### 3.2.1 The Proposed Baseline Network and Variants

Without extensively searching and refining the hyper-parameters, which could systematically underestimate the reported generalisation error, an empirically configured "Baseline" network was trained using the Adam optimiser starting at a learning rate of $10^{-5}$, with a minibatch of 4, four full-sized image-label quartets. The deformation regularisation weight was set to $\alpha = 0.5$ between the bending energy and the multiscale Dice, described in Section 2. The weight decay was not used. Initial number of channels for feature maps was set to $n_0=32$. All network parameters were assigned initial values using Xavier initialiser (Glorot and Bengio, 2010), except for the final displacement prediction layers to allow controlled initial outputs as discussed in Section 2.3. These convolutional kernel and bias parameters were initialised to zeros for the results reported in this paper. We refer to the network trained with these hyper-parameters as the "Baseline" network, for comparing with the networks using different hyper-parameters. Except for each of the hyper-parameter of comparison, these configurations were kept fixed in the following networks.

Two variants of the proposed "Baseline" network loss function are compared, training with 1) a multiscale cross-entropy, described in Section 2.2 ("Baseline-msCE"), instead of the multiscale Dice, or 2) replacing the bending energy with an average $L^2$-norm of the displacement gradients ("Baseline-$L^2$").

Although one of the advantages of the proposed label similarity measure in Eq. (5) is computational efficiency when required on-the-fly, pre-computing Gaussian filtered labels before training, may further accelerate training. Therefore, a baseline network using label maps pre-filtered at different scales ("Baseline-preFilt") was trained, while the Dice metrics were evaluated directly on the resampled multiscale label maps during training.

To validate the proposed network architecture, the "Baseline" network was trained with only the displacement $\delta_0$ predicted at the input image resolution level $s_0$, i.e. without displacement summands $\delta_{1-4}$ at resolution levels $s_{1-4}$ ("Baseline-$\delta_0$", illustrated in Fig.5b). This is similar to the "Local-Net" proposed in our preliminary work (Hu et al., 2018). Furthermore, previous work suggested that, regularised displacements predicted at finest level may not be necessary (Dosovitskiy et al., 2015). Therefore, the "Baseline" network was also trained with all the displacement summands except for the one at level $s_0$, that is a network with displacement summed over the outputs at levels $s_{1-4}$ ("Baseline-$\delta_{1-4}$", illustrated in Fig.5c). For both networks, the down- and up-sampling blocks remain the same.

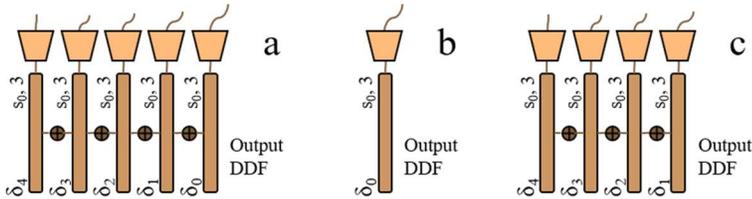

**Fig.5.** Illustration of the configuration variants for the output displacement summation used in the proposed baseline networks. **a** is adopted in the "Baseline" network using all the nodes $\delta_{0-4}$; **b** is in the "Baseline-$\delta_0$" network using only the prediction at the input image resolution level $s_0$; **c** represents the output configuration in the "Baseline-$\delta_{1-4}$" network without the prediction at the finest $s_0$ level.

#### 3.2.2 Comparison with the Previous Networks of (Hu et al., 2018)

A "Global-Net", illustrated in Fig.6, was proposed to predict an affine transformation using the same learning framework described in Section 2.1. A "Composite-Net" was proposed to compose the output DDFs from the "Global-Net" and the "Local-Net", as illustrated in Fig.7. The details of the compared "Global-Net" and the "Composite-Net", are described in (Hu et al., 2018). A direct comparison to the previously reported numerical results may be unfair due to the difference in data sets and the associated training strategy. For example, the results reported in this paper are based on substantially more anatomical labels verified by second observers (described in Section 3.1) without the less-frequently-sampled "low-confidence" labels (Hu et al., 2018). In the interest of a direct comparison between different network architectures, the "Global-Net" and the "Composite-Net" were re-trained using the same multiscale Dice as the "Baseline" networks, with a smaller starting learning rate of $10^{-6}$ to avoid otherwise frequently encountered divergence (due to the sensitivity of the output displacements to the affine parameters). A 24GB NVIDIA® Quadro™ P6000 GPU card was used to train the "Composite-Net" that needs more than 12GB GPU memory for the same minibatch size.

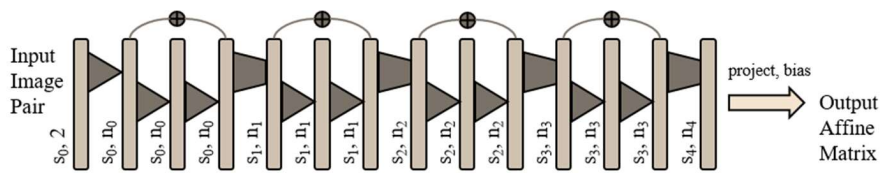

**Fig.6.** Illustration of the previously proposed "Global-Net". The "Global-Net" shares the same architecture (using independently learnable parameters) as the four down-sampling blocks of the "Local-Net". The details are described in (Hu et al., 2018).

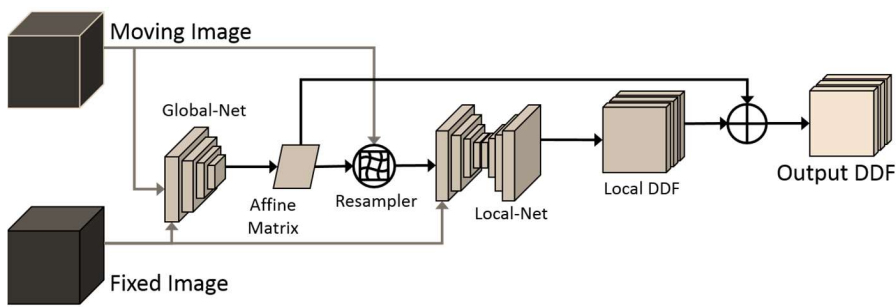

**Fig.7.** Illustration of the inference part of the previously proposed "Composite-Net", combining a "Global-Net" with a "Local-Net". The details are described in (Hu et al., 2018).

### 3.3 Cross-Validation

All the numerical results reported in this paper were based on a 12-fold patient-level cross-validation for each network. In each fold, test data from 6-7 patients were held out while the data from the remaining patients were used in training. Two measures are reported in this study: centroid distance error between centres of mass is computed from each pair of the warped and fixed labels; the target registration error (TRE) is defined as root-mean-square on these distance errors over all landmark pairs for each patient. A Dice similarity coefficient (DSC) is the overlap between the binary warped and fixed labels representing prostate glands. These two independently-calculated metrics on left-out test data directly relate to the clinical requirements in the registration-enabled guidance, avoiding surrounding healthy or vulnerable structures and locating regions of interest. Paired Wilcoxon signed-rank tests at significance level $\alpha_H$=0.05 were used to compare medians of the cross-validation results between the networks. Confidence intervals (CIs) were also reported in cases where the obtained p-values are larger than $\alpha_H$. The cross-validation scheme ensures all the anatomical landmarks (details described in Section 3.1) are independently tested in different folds without being used in training.

### 3.4 Comparison with Pairwise Image Registration Methods

As discussed in Section 1, generic pairwise registration algorithms were generally found to perform poorly in registering MR and TRUS images for this application, which has in turn motivated many application-specific methods, such as prostate motion modelling and intraoperative rigid initialisation, e.g. (De Silva et al., 2017). To confirm this observation on the same data set in this work, a set of non-linear registrations were tested using a GPU-enabled open-source algorithm (Modat et al., 2010). The B-splines free-form deformation regularised by bending energy (Rueckert et al., 1999), weighting being set to 0.5 for comparison, was optimised with respect to three intensity-based similarity measures, normalised mutual information (NMI), normalised cross-correlation (NCC) and sum-of-square differences (SSD). In addition to directly applying the registration without any initial alignment, two simple global initialisation methods, an automatic rigid registration minimising the same similarity measures and a manual initialisation matching the gland centroids, were also tested. A total of 972 registrations were run on GPU using the data set described in Section 3.1. The TREs and DSCs were computed with all the other default configurations kept as the same for comparison. These results aim to demonstrate typical performances using pairwise intensity-based registration algorithms for this multimodal MR-to-TRUS prostate imaging application. Methods with substantial customised adaptations (discussed in Section 1), such as spatial initialisation (manual or automated) or statistical motion modelling, were also compared quantitatively based on published results and are summarised in Section 4.4.

## 4 Results

### 4.1 "Baseline" Performance

Approximately four 3D registrations per second can be performed on the same GPUs. The "Baseline" network achieved a median TRE of 3.6 mm on landmark centroids with first and third quartiles being 2.3 and 6.5 mm, respectively. A median DSC of 0.87 on prostate glands was obtained from the same networks with first- and third quartiles being 0.82 and 0.89. More detailed results are summarised in Table 1 and illustrated in Fig.8. Example slices from the input MR and TRUS image pairs and the registered MR images are provided in Fig.9 for qualitative visual assessment of the registration results based on the test data.

### 4.2 Variants of the "Baseline" Network

Considering the "Baseline" network was trained with respect to the loss function based on multiscale Dice, it is interesting that replacing the multi-scale Dice with cross-entropy (i.e. using "Baseline-msCE" network) had a significantly worse TRE (*p-value<0.001*), but a better binary (single-scale) DSC result (*p-value=0.046*). This may suggest that the superior class balance was conveyed by the multiscale Dice as discussed in Section 2.2. Thus, the bias towards labels having larger volumes, such as the prostate glands producing the DSC results, is lessened. The "Baseline-$L^2$" using a different deformation regularisation produced poorer generalisation ability, both in terms of TRE (*p-value=0.049*) and DSC (with both *p-values<0.001*), although it is intended to demonstrate the suitability to use different forms of regularisation without excessively tuning each hyper-parameter in this experiment.

It may be of practical importance to report that pre-computing the label filtering did not have a negative impact on TRE (*p-value=0.458, CI=[-1.433, 0.634]*) or on DSC (*p-value=0.498, CI=[-0.009, 0.030]*). However, the "Baseline-preFilt" is faster to train. Depending on the implementation of the online filtering and the parsing of the additional pre-computed labels, an approximately 25% gain in training time was achieved in our experiments using pre-computed labels.

The "Baseline" network outperformed the "Baseline-$\delta_0$" network predicting the local displacement only at the finest input image resolution level, with *p-value=0.034* and *p-value=0.003*, for comparing TREs and DSCs, respectively. This improvement was consistently achieved during the experiments with different network hyper-parameters. On the other hand, the "Baseline-$\delta_{1-4}$" without predicting displacement at finest resolution level performed competitively, consistent with the conclusions from the previous work (Dosovitskiy et al., 2015) that prediction at the original resolution level does not necessarily improve the accuracy. It produced TREs and DSCs with no statistically significant difference than those from the "Baseline", *p-value=0.477 (CI=[-1.342, 0.735])* and *p-value=0.316 (with a CI of [-0.011, 0.023])*, respectively. Furthermore, using the "Baseline" network without the trilinear additive up-sampling layers, described in Section 2.3, resulted in a significantly higher median TRE of 6.4 mm (*p-value<0.001*).

### 4.3 Comparison Results with the Previous Networks of (Hu et al., 2018)

The TREs and DSCs from the "Baseline" network are significantly better than those from "Global-Net" which only models the affine transformation (both *p-values<0.001*). This clearly demonstrates the efficacy of the deformable registration in this application. Comparing to the previously proposed "Composite-Net" architecture, not only the GPU memory to train the "Global-Net" can be spared, but also improvement in generalisation was observed from the proposed network, in terms of both TRE and DSC (both *p-values<0.001*).

Because a relatively large weight $\alpha = 0.5$ in Eq. (2) was used in this multimodal application, negative Jacobian determinants were not found in any of the DDFs predicted by the trained networks. For further inspection of the deformation fields, we plotted the determinants of the Jacobian, the magnitudes of the displacement vectors and the $L^2$-norms of the displacement gradients, as illustrated in the rows of Fig.10, J, D and G, respectively. For example, the "Baseline" (left columns), "Baseline-$\delta_0$" (middle columns) and an illustrative network trained with small regularisation weight $\alpha = 0.01$ (right columns) produced DDFs with visibly increasing variance. Both standard deviations and numerical ranges of these three quantities increase in the same order consistently. Negative Jacobian determinants also appeared as the regularisation weight decreases to $\alpha = 0.01$, implying that physically implausible deformation may exist in the illustrative example without appropriate regularisation.

### 4.4 Comparison with Pairwise Registration Methods

For the comparison with the pairwise registrations described in Section 3.4, we report that all 9 median TREs are larger than 24 mm and none of the DSC medians are higher than 0.77. Direct application of the intensity-based registration result in median TREs ranging 26.7-35.0 mm, with and without the rigid initialisation, for all three similarity measures. Manually aligning the prostate gland centroids immediately led to a median TRE of 19.6 mm with a median DSC of 0.79, without further registration. With the manual centroid-alignment as initialisation, registrations using NMI, NCC and SSD produced higher median TREs of 20.6, 24.7 and 25.6 mm, with lower median DSCs of 0.77, 0.67 and 0.65, respectively. The results are also summarised and compared with other previously proposed methods in Table 2, with an initial median TRE of 34.8 mm before registration. These inferior performances appear much worse than the results summarised in Table 1 and those from previous application-specific methods, e.g. (De Silva et al., 2017; Hu et al., 2012; Khallaghi et al., 2015; Sun et al., 2015; van de Ven et al., 2015; Wang et al., 2016a). It should clearly indicate the nontrivial difficulties for these general-purpose intensity-based algorithms in this multimodal registration application.

For the same application, the previous studies validated on patient data reported an expected-TRE range of 1.4-2.8 mm, (De Silva et al., 2017; Hu et al., 2012; Khallaghi et al., 2015; van de Ven et al., 2015; Wang et al., 2016a). These results were based on smaller sample sizes (ranging from 8 to 29 cases) with significant variations, for example, an individual-TRE range of 0.8-8.0 mm (van de Ven et al., 2015) was reported. Although intensity-based registration has also been adopted for this application, they usually rely on customised optimisation and/or manual initialisation. For instance, a previous study (Sun et al., 2015) reported a median TRE of 1.8 mm on 20 patients, using a dual optimisation with modality independent neighbourhood descriptor after an initialisation method based on six manual landmarks from expert observers for each registration. Our method is fully automated without requiring any initialisation, pre- or intra-procedural segmentation, once the registration network is trained. One of the latest developments also reported an automated initialisation based on predicting rigid prostate motion (De Silva et al., 2017), but all the other approaches still require either manual (partial) segmentation of the TRUS images or manual initialisation in order to obtain robust registrations. None of these methods reported a faster registration execution time than the sub-second performance with the proposed registration network.

| Networks | TRE in mm | | DSC % | |
|---|---|---|---|---|
| | Median | Percentiles [$10^{th}$, $25^{th}$, $75^{th}$, $90^{th}$] | Median | Percentiles [$10^{th}$, $25^{th}$, $75^{th}$, $90^{th}$] |
| Baseline | 3.6 | [1.6, 2.3, 6.5, 10.0] | 0.87 | [0.77, 0.82, 0.89, 0.91] |
| Baseline-msCE | **6.1** | [1.8, 3.3, 9.0, 13.2] | **0.88** | [0.77, 0.83, 0.90, 0.93] |
| Baseline-$L^2$ | **4.8** | [1.7, 2.7, 7.7, 11.6] | **0.82** | [0.68, 0.76, 0.86, 0.88] |
| Baseline-preFilt | 3.9 | [1.6, 2.4, 7.0, 10.2] | 0.86 | [0.74, 0.81, 0.88, 0.90] |
| Baseline-$\delta_0$ | **4.5** | [1.9, 2.8, 7.5, 11.3] | **0.84** | [0.72, 0.79, 0.87, 0.89] |
| Baseline-$\delta_{1-4}$ | 4.2 | [1.5, 2.4, 6.6, 10.4] | 0.86 | [0.74, 0.82, 0.89, 0.90] |
| Global-Net | **5.8** | [2.3, 3.8, 8.6, 12.0] | **0.77** | [0.62, 0.69, 0.82, 0.84] |
| Composite-Net | **4.7** | [2.3, 3.3, 7.5, 10.5] | **0.82** | [0.68, 0.78, 0.86, 0.87] |

**Table 1.** Summary of the cross-validation results for the networks described in Section 3.2. The medians in bold numbers indicate statistically significant deviation from the "Baseline" network.

| Registration Method | Expected TRE in mm | No. of Cases | Initialisation Method |
|---|---|---|---|
| Initial | 34.8 (median) | 108 | n/a |
| After centroid-alignment | 19.6 (median) | 108 | n/a |
| FFD with NMI* | 20.6 (median) | 108 | Gland centroids |
| FFD with NCC* | 24.7 (median) | 108 | (from prostate gland/surface estimates) |
| FFD with SSD* | 25.6 (median) | 108 | |
| Hu 2012 | 2.4 (median) | 8 | Manual landmarks |
| Khallaghi 2015 | 2.4 (mean) | 19 | Gland centroids |
| van de Ven 2015 | 2.8 (median) | 10 | Rigid surface registration |
| Sun 2015 | 1.8 (median) | 20 | Manual landmarks |
| Wang 2016 | 1.4 (mean) | 18 | Rigid surface registration |
| De Silva 2017 | 2.3 (mean) | 29 | Learned motion model |

**Table 2.** Summary of the results from the intensity-based nonrigid image registrations and those from other previous studies, described in Section 4.4. *The registration results included here are from simplified experiments on our data. It reflects a baseline performance of the compared intensity-based methods without application-specific adaptation, such as initialisation method, registration parameters and other similarity measures.

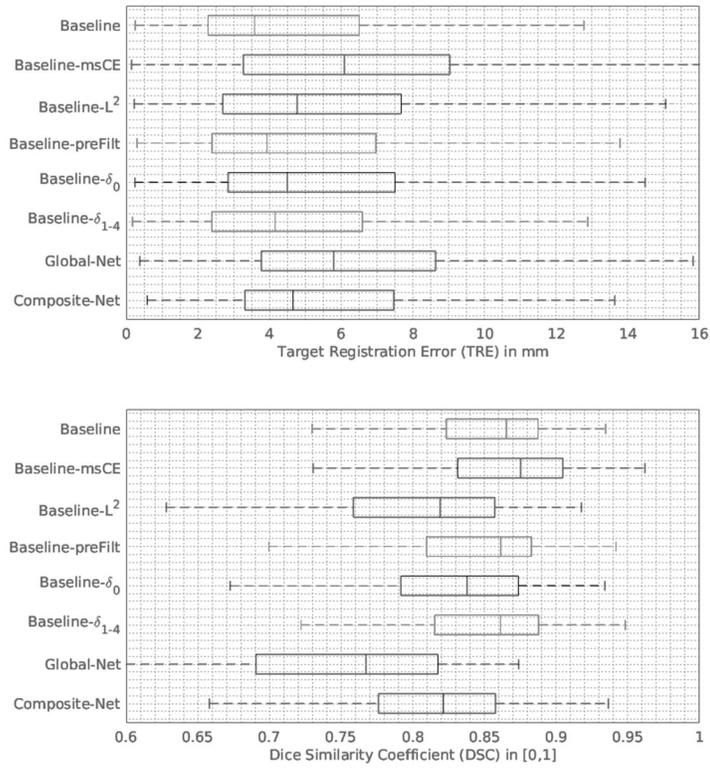

**Fig.8.** Tukey's boxplots of the cross-validation results obtained from the networks described in Section 3.2. The numerical results are also summarised in Table 1.

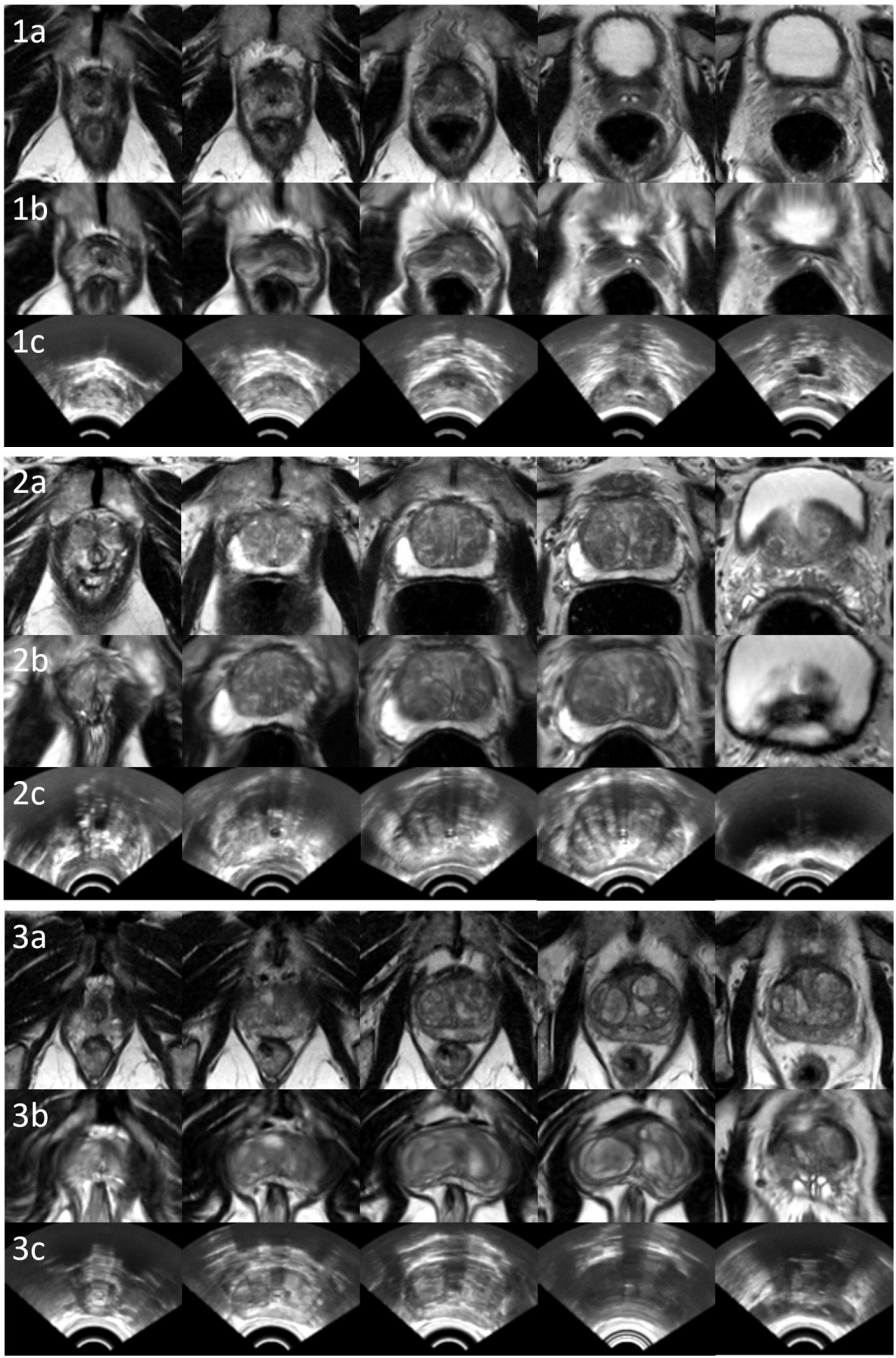

**Fig.9.** Example image slices from three test cases, 1, 2 and 3. Rows a, b and c contain slices from original MR images (visually closest slices chosen manually for comparison), equidistant slices from the warped moving MR images using the proposed "Baseline" network, and the corresponding fixed TRUS images, respectively.

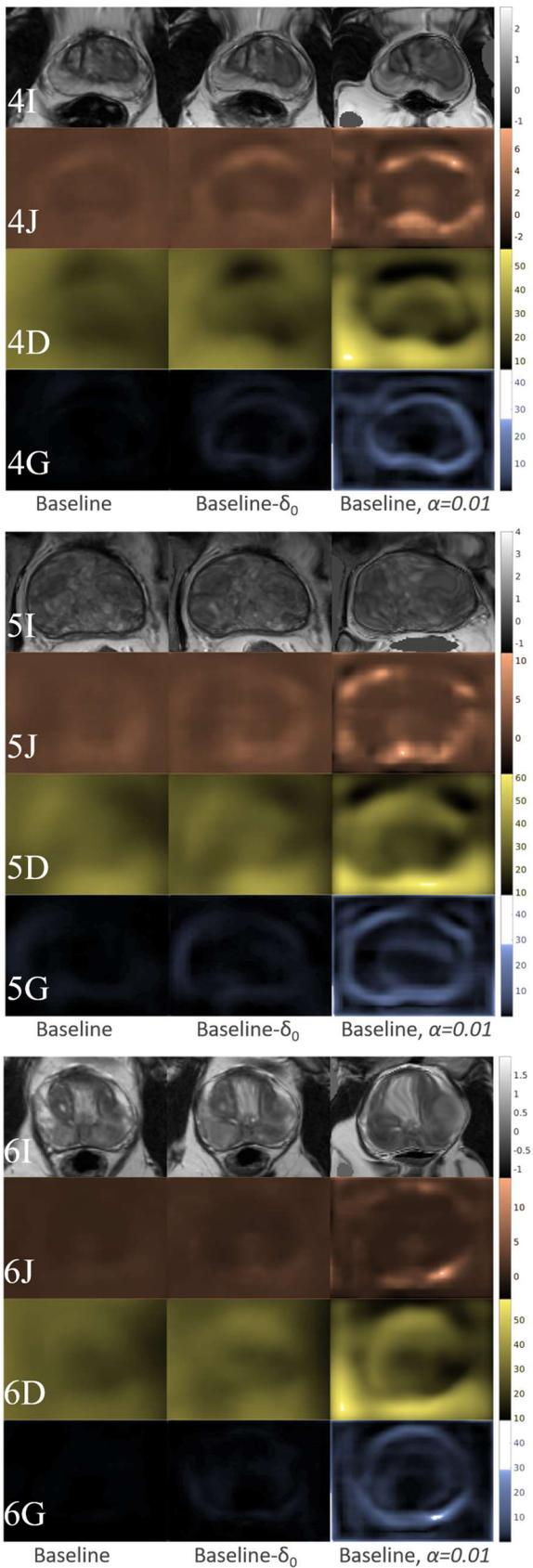

**Fig.10.** Inspection of the warped MR images and network-predicted DDFs from three test cases, 4, 5 and 6: The first (I) rows (grey-scaled) display the warped intensity images; the second (J) rows (orange-scaled) plotted the determinants of the Jacobian; the third- (D) and fourth (G) rows (yellowed- and blue-scaled) plotted the magnitudes of the displacement vectors and the $L^2$-norms of the displacement gradients, respectively. Three columns contain results from three networks, the "Baseline" (left column), "Baseline-δ0" (middle column) and an illustrative baseline network trained with small regularisation weight $\alpha = 0.01$, respectively.

## 5 Discussion

In this work, we demonstrated the feasibility of non-iterative prediction of voxel correspondence from unlabelled input images, using training image pairs with only sparse annotations. The proposed method targets a wide range of clinical applications, where automatic multimodal image registration has been traditionally challenging due to the lack of reliable image similarity measures or automatic landmark extraction methods.

The use of sparse *training* and *validation* labels to predict and evaluate dense correspondence raises interesting open questions. The sparse training landmark pairs cannot independently represent voxel-level dense correspondence for an individual case. This is commonly addressed by application-independent deformation smoothness penalty in pairwise methods. Our architecture enables the regularised DDF to be implicitly learned from samples of latent dense correspondences, with the presented results suggesting that the population-trained application-specific regularisation improves the registration accuracy on unseen landmarks. For validation of dense correspondence, in the absence of ground-truth correspondence maps for real patient data, using sparse landmarks has become standard practice, interpreting independent landmark misalignments as samples of the dense registration error, e.g. (De Silva et al., 2017; Hu et al., 2012; Khallaghi et al., 2015; van de Ven et al., 2015; Wang et al., 2016a). All these studies adopted the same validation strategy based on available anatomical landmarks within or around prostate glands (described in Section 3.1), which have been shown to represent a spatial distribution relevant to the clinical localisation and targeting applications. Although MR and TRUS prostate images have limited number of salient corresponding features (approximately eight landmark pairs per image were annotated in this work), pooling these samples across 108 cases has enabled us to measure sub-millimetre accuracy differences with statistical significance. In practice, reliably finding substantially more paired corresponding anatomies has been proven challenging for experienced clinicians and researchers. Therefore, it is our opinion that further improvement in registration performance in terms of more accurate prediction of voxel correspondence may resort to increasing number of image/subject pairs or better regularisation strategy containing prior knowledge of the application-specific deformation, rather than increasing the number of landmarks per image pair.

In this work, we propose the multiscale Dice in Eq. (5) because of its ability to balance the inter-class gradient difference, discussed in Section 2.2, although the cross-entropy loss has an arguably more interpretable probability formulation for the weak voxel-level correspondence (Hu et al., 2018). Methods with weighting strategies such as generalised Dice (Sudre et al., 2017) and weighted cross-entropy (Ronneberger et al., 2015) did not seem to further improve the results in our application, probably due to the highly constrained outputs in the registration task. It is also interesting that some training labels overlap with each other, such as the gland segmentations and those defined within the prostate glands. Further quantitative analysis may be interesting to reveal the effect of these overlaps on registration performance. We envisage that, instead of heuristic weight-balancing to improve performance metrics, future investigation shall focus on risk analysis (Elkan, 2001) for specific applications to quantitatively optimise the utilities of the registration, such as those associated with clinical risks.

The DDFs, also discussed in Section 4.3, were predicted without explicitly enforced topology preservation, due to the relatively heavy regularisation required in this application. However, in applications where larger numbers of landmarks can be identified feasibly and larger deformations are clinically plausible, the network may be adapted, e.g. to penalise Jacobian-based regulariser, in seeking highly accurate registration. Furthermore, the final displacement field in our proposed network could also be represented by a composition of outputs $\delta_{0-4}$, instead of the proposed summation. It is computationally more expensive and potentially more sensitive to learning rate and initialisation, but may predict meaningful DDF components at different resolution levels, for instance, for allowing multi-level sparsity regularisation (Schnabel et al., 2001; Shi et al., 2012).

Whilst the reported cross-validation results were based on independent landmarks unseen in training, we would like to note that a limitation in the validation is that a sizable data set completely unseen to the methodology development was not available to test the generalisation ability conclusively. This is why we resort to cross-validation and did not pursue exhaustive hyper-parameter tuning. For example, the weight of bending energy was fixed among the baseline networks but was only set empirically after a limited number of trial runs on partial data set. Unbiased model searching methods for small- to medium sized training data remain an interesting future research direction.

In summary, we have introduced a registration framework that is flexible enough to utilise different neural network architectures, deformation regularisers, and anatomical features with varied sizes, shapes and availabilities, and to match input image intensity patterns. The trained network enables a fast and fully-automatic multimodal image registration algorithm using only input image pair. Registration results are reported from a validation on 108 labelled intraoperative prostate image pairs. Future research aims to investigate the generalisation of the proposed method to data from different centres and to a wider range of applications.

Acknowledgment

Y. Hu was funded by the UK Engineering and Physical Sciences Research Council (EPSRC) and Cancer Research UK (CRUK) via a CMIC Platform Fellowship (EP/M020533/1) and the UCL-KCL Comprehensive Cancer Imaging Centre. This work also received additional support from the Wellcome Trust, CRUK and the EPSRC (C28070/A19985; WT101957; 203145Z/16/Z; NS/A000027/1; EP/N026993/1). The authors would like to thank colleagues Weidi Xie from Oxford and Carole Sudre from UCL for helpful discussions, and Rachael Rodell from SmartTarget Ltd. for the assistance in data collection.

References

Abadi, M., Agarwal, A., Barham, P., Brevdo, E., Chen, Z., Citro, C., Corrado, G.S., Davis, A., Dean, J., Devin, M., others, 2016. Tensorflow: Large-scale machine learning on heterogeneous distributed systems. arXiv Prepr. arXiv1603.04467.

Bengio, Y., Courville, A., Vincent, P., 2013. Representation learning: A review and new perspectives. IEEE Trans. Pattern Anal. Mach. Intell. 35, 1798–1828. https://doi.org/10.1109/TPAMI.2013.50

Cao, X., Yang, J., Zhang, J., Nie, D., Kim, M., Wang, Q., Shen, D., 2017. Deformable image registration based on similarity-steered CNN regression, in: Lecture Notes in Computer Science (Including Subseries Lecture Notes in Artificial Intelligence and Lecture Notes in Bioinformatics). pp. 300–308. https://doi.org/10.1007/978-3-319-66182-7_35

De Silva, T., Cool, D.W., Yuan, J., Romagnoli, C., Samarabandu, J., Fenster, A., Ward, A.D., 2017. Robust 2-D-3-D Registration Optimization for Motion Compensation during 3-D TRUS-Guided Biopsy Using Learned Prostate Motion Data. IEEE Trans. Med. Imaging 36, 2010–2020. https://doi.org/10.1109/TMI.2017.2703150

de Vos, B.D., Berendsen, F.F., Viergever, M.A., Staring, M., Išgum, I., 2017. End-to-End Unsupervised Deformable Image Registration with a Convolutional Neural Network. DLMIA 2017, ML-CDS 2017, Lect. Notes Comput. Sci. 10553, 204–212. https://doi.org/10.1007/978-3-319-67558-9_24

Dickinson, L., Ahmed, H.U., Allen, C., Barentsz, J.O., Carey, B., Futterer, J.J., Heijmink, S.W., Hoskin, P.J., Kirkham, A., Padhani, A.R., Persad, R., Puech, P., Punwani, S., Sohaib, A.S., Tombal, B., Villers, A., Van Der Meulen, J., Emberton, M., 2011. Magnetic resonance imaging for the detection, localisation, and characterisation of prostate cancer: Recommendations from a European consensus meeting. Eur. Urol. 59, 477–494. https://doi.org/10.1016/j.eururo.2010.12.009

Donaldson, I., Hamid, S., Barratt, D., Hu, Y., Rodell, R., Villarini, B., Bonmati, E., Martin, P., Hawkes, D., McCartan, N., others, 2017. MP33-20 The smarttarget biopsy trial: a prospective paired blinded trial with randomisation to compare visual-estimation and image-fusion targeted prostate biopsies. J. Urol. 197, e425.

Dosovitskiy, A., Fischery, P., Ilg, E., Hausser, P., Hazirbas, C., Golkov, V., Smagt, P. Van Der, Cremers, D., Brox, T., 2015. FlowNet: Learning optical flow with convolutional networks, in: Proceedings of the IEEE International Conference on Computer Vision. pp. 2758–2766. https://doi.org/10.1109/ICCV.2015.316

Dou, Q., Yu, L., Chen, H., Jin, Y., Yang, X., Qin, J., Heng, P.A., 2017. 3D deeply supervised network for automated segmentation of volumetric medical images. Med. Image Anal. 41, 40–54. https://doi.org/10.1016/j.media.2017.05.001

Elkan, C., 2001. The foundations of cost-sensitive learning, in: IJCAI International Joint Conference on Artificial Intelligence. pp. 973–978. https://doi.org/doi=10.1.1.29.514

Ethan J. Halpern, 2008. Urogenital Ultrasound: A Text Atlas, 2nd ed. Radiology 248, 390–390. https://doi.org/10.1148/radiol.2482082516

Fischer, B., Modersitzki, J., 2004. A unified approach to fast image registration and a new curvature based registration technique, in: Linear Algebra and Its Applications. pp. 107–124. https://doi.org/10.1016/j.laa.2003.10.021

Garcia-Peraza-Herrera, L.C., Li, W., Fidon, L., Gruijthuijsen, C., Devreker, A., Attilakos, G., Deprest, J., Poorten, E. Vander, Stoyanov, D., Vercauteren, T., others, 2017. Toolnet: Holistically-nested real-time segmentation of robotic surgical tools. arXiv Prepr. arXiv1706.08126.

Ghavami, N., Hu, Y., Bonmati, E., Rodell, R., Gibson, E., Moore, C.M., Barratt, D.C., 2018. Automatic slice segmentation of intraoperative transrectal ultrasound images using convolutional neural networks, in: SPIE Medical Imaging.

Gibson, E., Giganti, F., Hu, Y., Bonmati, E., Bandula, S., Gurusamy, K., Davidson, B.R., Pereira, S.P., Clarkson, M.J., Barratt, D.C., 2017a. Towards image-guided pancreas and biliary endoscopy: Automatic multi-organ segmentation on abdominal CT with dense dilated networks, Lecture Notes in Computer Science (including subseries Lecture Notes in Artificial Intelligence and Lecture Notes in Bioinformatics). https://doi.org/10.1007/978-3-319-66182-7_83

Gibson, E., Li, W., Sudre, C., Fidon, L., Shakir, D., Wang, G., Eaton-Rosen, Z., Gray, R., Doel, T., Hu, Y., others, 2017b. NiftyNet: a deep-learning platform for medical imaging. arXiv Prepr. arXiv1709.03485.

Glorot, X., Bengio, Y., 2010. Understanding the difficulty of training deep feedforward neural networks. PMLR 9, 249–256. https://doi.org/10.1.1.207.2059

Goodfellow, I., Bengio, Y., Courville, A., 2016. Deep Learning--book. MIT Press 521, 800. https://doi.org/10.1038/nmeth.3707

He, K., Zhang, X., Ren, S., Sun, J., 2016. Deep Residual Learning for Image Recognition, in: 2016 IEEE Conference on Computer Vision and Pattern Recognition (CVPR). pp. 770–778. https://doi.org/10.1109/CVPR.2016.90

Hill, D.L., Batchelor, P.G., Holden, M., Hawkes, D.J., 2001. Medical image registration. Phys. Med. Biol. 46, R1–R45. https://doi.org/10.1088/0031-9155/46/3/201

Hu, Y., 2013. Registration of magnetic resonance and ultrasound images for guiding prostate cancer interventions. UCL (University College London).

Hu, Y., Ahmed, H.U., Taylor, Z., Allen, C., Emberton, M., Hawkes, D., Barratt, D., 2012. MR to ultrasound registration for image-guided prostate interventions. Med. Image Anal. 16, 687–703. https://doi.org/10.1016/j.media.2010.11.003

Hu, Y., Carter, T.J., Ahmed, H.U., Emberton, M., Allen, C., Hawkes, D.J., Barratt, D.C., 2011. Modelling prostate motion for data fusion during image-guided interventions. Med. Imaging, … 30, 1887–1900. https://doi.org/10.1109/TMI.2011.2158235

Hu, Y., Gibson, E., Ahmed, H.U., Moore, C.M., Emberton, M., Barratt, D.C., 2015. Population-based prediction of subject-specific prostate deformation for MR-to-ultrasound image registration. Med. Image Anal. 26, 332–344. https://doi.org/10.1016/j.media.2015.10.006

Hu, Y., Kasivisvanathan, V., Simmons, L.A.M., Clarkson, M.J., Thompson, S.A., Shah, T.T., Ahmed, H.U., Punwani, S., Hawkes, D.J., Emberton, M., Moore, C.M., Barratt, D.C., 2017. Development and Phantom Validation of a 3-D-Ultrasound-Guided System for Targeting MRI-Visible Lesions during Transrectal Prostate Biopsy. IEEE Trans. Biomed. Eng. 64. https://doi.org/10.1109/TBME.2016.2582734

Hu, Y., Modat, M., Gibson, E., Ghavami, N., Bonmati, E., Moore, C.M., Emberton, M., Noble, J.A., Barratt, D.C., Vercauteren, T., 2018. Label-driven weakly-supervised learning for multimodal deformable image registration. Biomed. Imaging (ISBI), 2018 IEEE 15th Int. Symp.

Huang, G., Liu, Z., Weinberger, K.Q., van der Maaten, L., 2016. Densely connected convolutional networks. arXiv Prepr. arXiv1608.06993.

Ilg, E., Mayer, N., Saikia, T., Keuper, M., Dosovitskiy, A., Brox, T., 2017. FlowNet 2.0: Evolution of Optical Flow Estimation with Deep Networks. CVPR. https://doi.org/10.1109/CVPR.2017.179

Jaderberg, M., 2015. Spatial Transformer Networks. arXiv. https://doi.org/10.1038/nbt.3343

Khallaghi, S., Sanchez, C.A., Rasoulian, A., Nouranian, S., Romagnoli, C., Abdi, H., Chang, S.D., Black, P.C., Goldenberg, L., Morris, W.J., Spadinger, I., Fenster, A., Ward, A., Fels, S., Abolmaesumi, P., 2015. Statistical Biomechanical Surface Registration: Application to MR-TRUS Fusion for Prostate Interventions. IEEE Trans. Med. Imaging 34, 2535–2549. https://doi.org/10.1109/TMI.2015.2443978

Krebs, J., Mansi, T., Delingette, H., Zhang, L., Ghesu, F.C., Miao, S., Maier, A.K., Ayache, N., Liao, R., Kamen, A., 2017. Robust non-rigid registration through agent-based action learning, in: Lecture Notes in Computer Science (Including Subseries Lecture Notes in Artificial Intelligence and Lecture Notes in Bioinformatics). pp. 344–352. https://doi.org/10.1007/978-3-319-66182-7_40

Kumar, M., Dass, S., 2009. A total variation-based algorithm for pixel-level image fusion. IEEE Trans. Image Process. 18, 2137–2143. https://doi.org/10.1109/TIP.2009.2025006

Lawrence, S., Burns, I., Back, A., Tsoi, A.C., Giles, C.L., 2012. Neural network classification and prior class probabilities. Lect. Notes Comput. Sci. (including Subser. Lect. Notes Artif. Intell. Lect. Notes Bioinformatics) 7700 LECTU, 295–309. https://doi.org/10.1007/978-3-642-35289-8-19

LeCun, Y., Bengio, Y., Hinton, G., 2015. Deep learning. Nature. https://doi.org/10.1038/nature14539

LeCun, Y., Bottou, L., Bengio, Y., Haffner, P., 1998. Gradient-based learning applied to document recognition. Proc. IEEE 86, 2278–2323. https://doi.org/10.1109/5.726791

Lee, C.-Y., Xie, S., Gallagher, P., Zhang, Z., Tu, Z., 2015. Deeply-supervised nets, in: Artificial Intelligence and Statistics. pp. 562–570.

Liao, R., Miao, S., de Tournemire, P., Grbic, S., Kamen, A., Mansi, T., Comaniciu, D., 2017. An Artificial Agent for Robust Image Registration., in: AAAI. pp. 4168--4175.

Miao, S., Wang, Z.J., Liao, R., 2016. A CNN Regression Approach for Real-Time 2D/3D Registration. IEEE Trans. Med. Imaging 35, 1352–1363. https://doi.org/10.1109/TMI.2016.2521800

Milletari, F., Navab, N., Ahmadi, S.A., 2016. V-Net: Fully convolutional neural networks for volumetric medical image segmentation, in: Proceedings - 2016 4th International Conference on 3D Vision, 3DV 2016. pp. 565–571. https://doi.org/10.1109/3DV.2016.79

Modat, M., Ridgway, G.R., Taylor, Z.A., Lehmann, M., Barnes, J., Hawkes, D.J., Fox, N.C., Ourselin, S., 2010. Fast free-form deformation using graphics processing units. Comput. Methods Programs Biomed. 98, 278–284. https://doi.org/10.1016/j.cmpb.2009.09.002

Noble, J.A., 2016. Reflections on ultrasound image analysis. Med. Image Anal. 33, 33–37. https://doi.org/10.1016/j.media.2016.06.015

Panagiotaki, E., Chan, R.W., Dikaios, N., Ahmed, H.U., O'Callaghan, J., Freeman, A., Atkinson, D., Punwani, S., Hawkes, D.J., Alexander, D.C., 2015. Microstructural Characterization of Normal and Malignant Human Prostate Tissue With Vascular, Extracellular, and Restricted Diffusion for Cytometry in Tumours Magnetic Resonance Imaging. Invest. Radiol. 50, 218–227.

https://doi.org/10.1097/RLI.0000000000000115

Pereyra, G., Tucker, G., Chorowski, J., Kaiser, Ł., Hinton, G., 2017. Regularizing neural networks by penalizing confident output distributions. arXiv Prepr. arXiv1701.06548.

Pinto, P.A., Chung, P.H., Rastinehad, A.R., Baccala, A.A., Kruecker, J., Benjamin, C.J., Xu, S., Yan, P., Kadoury, S., Chua, C., Locklin, J.K., Turkbey, B., Shih, J.H., Gates, S.P., Buckner, C., Bratslavsky, G., Linehan, W.M., Glossop, N.D., Choyke, P.L., Wood, B.J., 2011. Magnetic resonance imaging/ultrasound fusion guided prostate biopsy improves cancer detection following transrectal ultrasound biopsy and correlates with multiparametric magnetic resonance imaging. J. Urol. 186, 1281–1285. https://doi.org/10.1016/j.juro.2011.05.078

Rastinehad, A.R., Turkbey, B., Salami, S.S., Yaskiv, O., George, A.K., Fakhoury, M., Beecher, K., Vira, M.A., Kavoussi, L.R., Siegel, D.N., Villani, R., Ben-Levi, E., 2014. Improving detection of clinically significant prostate cancer: Magnetic resonance imaging/transrectal ultrasound fusion guided prostate biopsy. J. Urol. 191, 1749–1754. https://doi.org/10.1016/j.juro.2013.12.007

Rohé, M.-M., Datar, M., Heimann, T., Sermesant, M., Pennec, X., 2017. SVF-Net: learning deformable image registration using shape matching, Lecture Notes in Computer Science (including subseries Lecture Notes in Artificial Intelligence and Lecture Notes in Bioinformatics). https://doi.org/10.1007/978-3-319-66182-7_31

Ronneberger, O., Fischer, P., Brox, T., 2015. U-Net: Convolutional Networks for Biomedical Image Segmentation. Med. Image Comput. Comput. Interv. -- MICCAI 2015 234–241. https://doi.org/10.1007/978-3-319-24574-4_28

Rueckert, D., Sonoda, L.I., Hayes, C., Hill, D.L., Leach, M.O., Hawkes, D.J., 1999. Nonrigid registration using free-form deformations: application to breast MR images. IEEE Trans. Med. Imaging 18, 712–21. https://doi.org/10.1109/42.796284

Schnabel, J.A., Rueckert, D., Quist, M., Blackall, J.M., Castellano-Smith, A.D., Hartkens, T., Penney, G.P., Hall, W.A., Liu, H., Truwit, C.L., Gerritsen, F.A., Hill, D.L.G., Hawkes, D.J., 2001. A generic framework for non-rigid registration based on non-uniform multi-level free-form deformations, in: Lecture Notes in Computer Science (Including Subseries Lecture Notes in Artificial Intelligence and Lecture Notes in Bioinformatics). pp. 573–581. https://doi.org/10.1007/3-540-45468-3_69

Shi, W., Zhuang, X., Pizarro, L., Bai, W., Wang, H., Tung, K.-P., Edwards, P., Rueckert, D., 2012. Registration using sparse free-form deformations. Med. Image Comput. Comput. Assist. Interv. 15, 659–66. https://doi.org/10.1007/978-3-642-33418-4_81

Siddiqui, M.M., Rais-Bahrami, S., Turkbey, B., George, A.K., Rothwax, J., Shakir, N., Okoro, C., Raskolnikov, D., Parnes, H.L., Linehan, W.M., Merino, M.J., Simon, R.M., Choyke, P.L., Wood, B.J., Pinto, P.A., 2015. Comparison of MR/ultrasound fusion-guided biopsy with ultrasound-guided biopsy for the diagnosis of prostate cancer. Jama 313, 390–7. https://doi.org/10.1001/jama.2014.17942

Simonovsky, M., Gutiérrez-Becker, B., Mateus, D., Navab, N., Komodakis, N., 2016. A deep metric for multimodal registration, in: Lecture Notes in Computer Science (Including Subseries Lecture Notes in Artificial Intelligence and Lecture Notes in Bioinformatics). pp. 10–18. https://doi.org/10.1007/978-3-319-46726-9_2

SmartTarget: BIOPSY [WWW Document], 2015. URL https://clinicaltrials.gov/ct2/show/NCT02341677

SmartTarget THERAPY [WWW Document], 2014. URL https://clinicaltrials.gov/ct2/show/NCT02290561

Sokooti, H., de Vos, B., Berendsen, F., Lelieveldt, B.P.F., Išgum, I., Staring, M., 2017. Nonrigid Image Registration Using Multi-scale 3D Convolutional Neural Networks, in: Medical Image Computing and Computer Assisted Intervention – MICCAI 2017: 20th International Conference, Quebec City, QC, Canada, September 11-13, 2017, Proceedings, Part I. pp. 232–239. https://doi.org/10.1007/978-3-319-66182-7_27

Sudre, C.H., Li, W., Vercauteren, T., Ourselin, S., Jorge Cardoso, M., 2017. Generalised dice overlap as a deep learning loss function for highly unbalanced segmentations, in: Lecture Notes in Computer Science (Including Subseries Lecture Notes in Artificial Intelligence and Lecture Notes in Bioinformatics). pp. 240–248. https://doi.org/10.1007/978-3-319-67558-9_28

Sun, Y., Yuan, J., Qiu, W., Rajchl, M., Romagnoli, C., Fenster, A., 2015. Three-dimensional nonrigid MR-TRUS registration using dual optimization. IEEE Trans. Med. Imaging. https://doi.org/10.1109/TMI.2014.2375207

Szegedy, C., Liu, W., Jia, Y., Sermanet, P., Reed, S., Anguelov, D., Erhan, D., Vanhoucke, V., Rabinovich, A., 2015. Going deeper with convolutions, in: Proceedings of the IEEE Computer Society Conference on Computer Vision and Pattern Recognition. pp. 1–9. https://doi.org/10.1109/CVPR.2015.7298594

Szegedy, C., Vanhoucke, V., Ioffe, S., Shlens, J., Wojna, Z., 2016. Rethinking the inception architecture for computer vision, in: Proceedings of the IEEE Conference on Computer Vision and Pattern Recognition. pp. 2818–2826.

Valerio, M., Donaldson, I., Emberton, M., Ehdaie, B., Hadaschik, B.A., Marks, L.S., Mozer, P., Rastinehad, A.R., Ahmed, H.U., 2015. Detection of clinically significant prostate cancer using magnetic resonance imaging-ultrasound fusion targeted biopsy: A systematic review. Eur. Urol. https://doi.org/10.1016/j.eururo.2014.10.026

van de Ven, W.J.M., Hu, Y., Barentsz, J.O., Karssemeijer, N., Barratt, D., Huisman, H.J., 2015. Biomechanical modeling constrained surface-based image registration for prostate MR guided TRUS biopsy. Med. Phys. 42, 2470–81. https://doi.org/10.1118/1.4917481

van de Ven, W.J.M., Hu, Y., Barentsz, J.O., Karssemeijer, N., Barratt, D., Huisman, H.J., 2013. Surface-based prostate registration with biomechanical regularization, in: SPIE Medical Imaging. p. 86711R--86711R.

Vargas, H.A., Hötker, A.M., Goldman, D.A., Moskowitz, C.S., Gondo, T., Matsumoto, K., Ehdaie, B., Woo, S., Fine, S.W., Reuter, V.E., Sala, E., Hricak, H., 2016. Updated prostate imaging reporting and data system (PIRADS v2) recommendations for the detection of clinically significant prostate cancer using multiparametric MRI: critical evaluation using whole-mount pathology as standard of reference. Eur. Radiol. 26, 1606–1612. https://doi.org/10.1007/s00330-015-4015-6

Viergever, M.A., Maintz, J.B.A., Klein, S., Murphy, K., Staring, M., Pluim, J.P.W., 2016. A survey of medical image registration. Med. Image Anal. 33, 140–144. https://doi.org/16/S1361-8415(01)80026-8

Vishnevskiy, V., Gass, T., Szekely, G., Tanner, C., Goksel, O., 2017. Isotropic Total Variation Regularization of Displacements in Parametric Image Registration. IEEE Trans. Med. Imaging 36, 385–395. https://doi.org/10.1109/TMI.2016.2610583

Wang, G., Li, W., Zuluaga, M.A., Pratt, R., Patel, P.A., Aertsen, M., Doel, T., David, A.L., Deprest, J., Ourselin, S., Vercauteren, T., 2017. Interactive medical image segmentation using deep learning with image-specific fine-tuning. arXiv Prepr. arXiv1710.04043.

Wang, Y., Cheng, J.Z., Ni, D., Lin, M., Qin, J., Luo, X., Xu, M., Xie, X., Heng, P.A., 2016a. Towards personalized statistical deformable model and hybrid point matching for robust MR-TRUS registration. IEEE Trans. Med. Imaging 35, 589–604. https://doi.org/10.1109/TMI.2015.2485299

Wang, Y., Ni, D., Qin, J., Xu, M., Xie, X., Heng, P.A., 2016b. Patient-specific Deformation Modelling via Elastography: Application to Image-guided Prostate Interventions. Sci. Rep. 6. https://doi.org/10.1038/srep27386

Wilson, D.M., Kurhanewicz, J., 2014. Hyperpolarized 13C MR for Molecular Imaging of Prostate Cancer. J. Nucl. Med. 55, 1567–1572. https://doi.org/10.2967/jnumed.114.141705

Wojna, Z., Ferrari, V., Guadarrama, S., Silberman, N., Chen, L.-C., Fathi, A., Uijlings, J., 2017. The Devil is in the Decoder. arXiv Prepr. arXiv1707.05847.

Wu, G., Kim, M., Wang, Q., Gao, Y., Liao, S., Shen, D., 2013. Unsupervised deep feature learning for deformable registration\nof MR brain images. Med. Image Comput. Comput. Assist. Interv. 16, 649–656.

Yang, X., Kwitt, R., Styner, M., Niethammer, M., 2017. Quicksilver: Fast predictive image registration – A deep learning approach. Neuroimage 158, 378–396. https://doi.org/10.1016/j.neuroimage.2017.07.008

Yu, J.J., Harley, A.W., Derpanis, K.G., 2016. Back to basics: Unsupervised learning of optical flow via brightness constancy and motion smoothness, in: Lecture Notes in Computer Science (Including Subseries Lecture Notes in Artificial Intelligence and Lecture Notes in Bioinformatics). pp. 3–10. https://doi.org/10.1007/978-3-319-49409-8_1

Yu, L., Yang, X., Chen, H., Qin, J., Heng, P.-A., 2017. Volumetric ConvNets with Mixed Residual Connections for Automated Prostate Segmentation from 3D MR Images. Thirty-First AAAI Conf. Artif. Intell. 66–72.

Zöllei, L., Fisher, J.W., Wells, W.M., 2003. A unified statistical and information theoretic framework for multi-modal image registration. Inf. Process. Med. Imaging 18, 366–77.